\UseRawInputEncoding
\documentclass[conference]{IEEEtran}
\IEEEoverridecommandlockouts

\usepackage{cite}
\usepackage{amsmath,amssymb,amsfonts}
\usepackage{algorithmic}
\usepackage{graphicx}
\usepackage{subcaption}
\usepackage{textcomp}
\usepackage{booktabs}
\usepackage{xcolor}
\usepackage{hyperref}

\def\BibTeX{{\rm B\kern-.05em{\sc i\kern-.025em b}\kern-.08em
    T\kern-.1667em\lower.7ex\hbox{E}\kern-.125emX}}
\begin{document}

\title{PPDL: A Real-world Industrial User Retention Ratio Forecasting Framework Integrating Physical Priors with Deep Learning}


\author{\IEEEauthorblockN{Zibo Zhao\textsuperscript{*,1}, Zhengxiong Guan\textsuperscript{*,2}, Chaoli Zhang\textsuperscript{\textdagger,1}, Linyuan Geng\textsuperscript{2}, Xuanbing Zhu\textsuperscript{2}, Zhonglong Zheng\textsuperscript{\textdagger,1}, Fan Wu\textsuperscript{3}}
\IEEEauthorblockA{\textsuperscript{1}\textit{Zhejiang Normal University}, \textsuperscript{2}\textit{Douyin Group}, \textsuperscript{3}\textit{Shanghai Jiao Tong University}  \\
\{zibozhao, chaolizcl, zhonglong\}@zjnu.edu.cn, \{guanzhengxiong, genglinyuan, zhuxuanbing\}@bytedance.com, fwu@cs.sjtu.edu.cn}
}


\maketitle

\let\thefootnoteOld\thefootnote
\renewcommand{\thefootnote}{\textasteriskcentered}
\footnotetext{These authors contributed equally to this work.}
\renewcommand{\thefootnote}{\textdagger}
\footnotetext{Corresponding authors.}
\renewcommand{\thefootnote}{}
\footnotetext{This work is supported in part by the NSFC under Grant No. 62502456; in part by the Zhejiang Provincial Natural Science Foundation of China under Grant No. LQN25F020020; in part by Open Research Fund of Zhejiang Key Laboratory of Intelligent Education Technology and Application under Grant No. 2025ZNJYKF013; in part by the Major Program of the Natural Science Foundation of Zhejiang Province under Grant No. LD26F020003; in part by the NSFC under Grant No. 62672453.}
\let\thefootnote\thefootnoteOld

\begin{abstract}
In multi-channel paid user acquisition, early and accurate prediction of user retention at the channel level is crucial for optimizing budget allocation. 
User retention curves display a pronounced temporal pattern: an initial period of high churn transitions into long-term stability. This pattern is further characterized by regular fluctuations attributable to seasonality and exhibits high serial autocorrelation. These intrinsic properties make such curves highly suitable for analysis within a time-series forecasting framework. However, forecasting user retention ratio for large-scale short-video platform faces three major challenges: significant heterogeneity across channels, pronounced global trend of decay followed by saturation, and short look-back windows. To address these challenges, we propose \textbf{PPDL}, a novel forecasting framework that integrates physical priors with deep learning. 
We first introduce a trend-residual decomposition component. The trend is modeled using the Weibull distribution, whose parameters are learned via a Multilayer Perceptron (MLP). Secondly, for the residual component, we design an auxiliary embedding module on top of a deep learning backbone to maintain the channel identity awareness. 
Finally, to enhance the model's sensitivity to trends, we design a Multiscale Trend-penalized loss function. The proposed approach PPDL is validated through comprehensive experiments on industrial-scale datasets, covering three applications with an average of 30+ channels each. Experimental results show that PPDL achieves improvements across different backbones and significantly outperforms existing online solutions. 
\end{abstract}


\section{Introduction}
\begin{figure}[t]
    \centering
    \includegraphics[width=1\columnwidth]{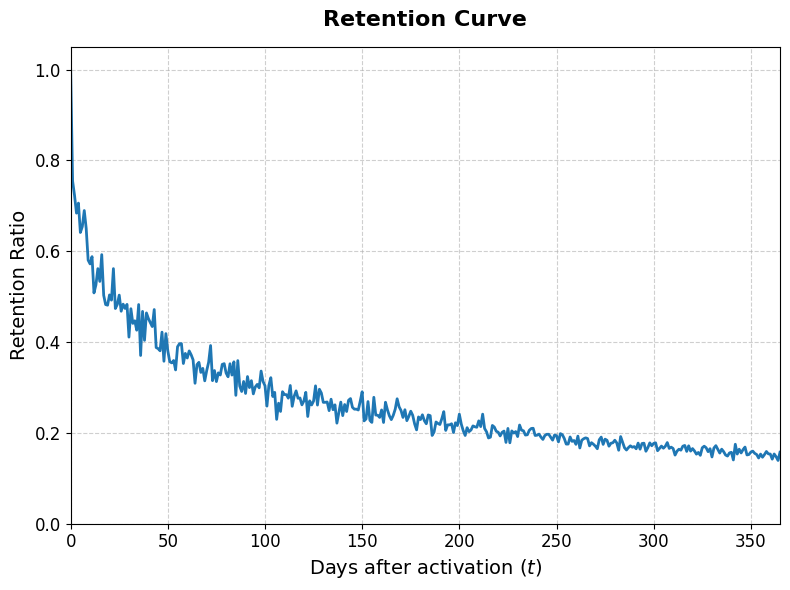}
    \setlength{\abovecaptionskip}{0.cm}
    \caption{
    The retention ratio $R(t)$ denotes the fraction of cohort users remaining active in $t$ days after activation. We call this time series of retention ratios the retention curve. }
    \label{Figure 1}
\end{figure}

For short-video platforms~\cite{cai2023reinforcing}, user stickiness is crucial to business growth and is typically measured by user retention ratio. 
As shown in Figure~\ref{Figure 1}, the retention ratio $R(t)$ denotes the fraction of cohort users remaining active in $t$ days after activation. 
To quantify the overall retention performance of a cohort, we aggregate the retention ratios by computing their cumulative sum. We name this metric as the cohort's LT (LifeTime), denoted 
$\text{LT}_n = \sum_{t=0}^{n} R(t)$. $\text{LT}_{n}$ 
provides a comprehensive measure of the quality of the channel over $n$ days. 
Considering product return cycles, $n$ is typically set as 365. To optimize budget allocation, we need to predict $\text{LT}_{365}$ at day 30 after activation. 
That is, the retention ratio of the following 335 days should be predicted given 30-day observed. The user retention curve shown in Figure~\ref{Figure 1} exhibits a pattern of gradual decay over time, accompanied by periodicity-like regular fluctuations. This temporal pattern is suitable for discussion within a time series forecasting framework.

In this scenario, user retention ratio forecasting primarily faces three challenges: channel heterogeneity, pronounced global decay-then-saturation trend, and short look-back window. Firstly, different channels exhibit significant differences due to factors such as user activation volume, channel-specific promotions and so on, 
resulting in retention curves that fluctuate intensely.
Secondly, user retention curves exhibit pronounced global decay-then-saturation trend, where periodic information can be obscured, leading to a potential loss of fine-grained information. 
Thirdly, since early and accurate prediction of user retention ratio is crucial for subsequent advertising decisions, the model is required to forecast user retention ratio over 335 time steps based on merely 30 days of observed data. 

Existing research~\cite{zhang2021user,tamilkodi2024unraveling,ullah2019churn} on user retention and churn prediction primarily focuses on the user-level. These approaches estimate user-level retention probability based on the machine learning model.
However, in our scenario, the behavioral data of newly acquired users is sparse, making it difficult to rely on user characteristics for user-level forecasting. We focus on channel-level forecasting.
Existing channel-level forecasting approaches~\cite{cherkashin2024practical} primarily employ statistical methods. Such static parametric approaches struggle to capture cross-channel dependencies, resulting in limited accuracy in modeling temporal patterns when confronted with heterogeneous channels. 

In statistical terms, according to the Fisher–Tippett–Gnedenko theorem~\cite{eaebbc4c-5f6a-3b31-9c38-cfb26bccff39} from extreme value theory, the retention curve consistently conforms to the Weibull distribution~\cite{rinne2008weibull} pattern. See Section~\ref{Preliminaries} for the detailed description. However, high variance among different channels 
makes parametric fitting challenging. What's more, it is difficult for a static parametric approach to capture dynamic factors in retention curve, such as seasonal effects, competitive pressures, holiday impact and so on. 
In contrast, time series forecasting models excel at processing multi-channel information, capturing cross-channel dependencies and similar temporal patterns to achieve accurate predictions. Therefore, we need to integrate channel-specific Weibull distribution with deep learning-based time series forecasting models for complex feature representation to forecast retention curve in real-world scenario. 

To address the challenges mentioned above, we propose a framework integrating physical priors with deep learning model called \textbf{PPDL}. PPDL consists of three key components. First, we implement the Weibull-Prior trend extractor. 
We model the trend through the Weibull distribution, and estimate its parameters via a gated-MLP based on covariates in scenario. Second, for the residual component, we incorporate auxiliary embedding on an MLP-based time series forecasting backbone to inspire channel identification and phase information representation. 
Third, to enhance the model’s ability to learn trends, 
we propose a multiscale trend-penalized loss function. To the best of our knowledge, our approach is the first to employ deep learning-based time series forecasting for channel-level user retention ratio forecasting. Our core contributions are as follows:

\begin{itemize}
    \item We employ trend-residual decomposition and novelly introduce Weibull distribution parameterized via MLP to model the trend information based on the characteristics of the user retention curves.
    \item Based on the deep learning backbone, we introduce an auxiliary embedding mechanism designed to provide channel identity and phase information for residual component prediction. Meanwhile, we propose a trend-penalized loss function to accurately guide the model in learning data trends, thereby enhancing prediction accuracy.
    \item The experimental results demonstrate that our framework outperforms the basic backbone model and the current online version.
\end{itemize}

\section{Preliminaries}
\label{Preliminaries}

This section systematically elaborates the core concepts and notation system for user retention ratio forecasting in paid acquisition.


\textbf{Definition 1 (Activation Date)} \textit{Activation date refers to the specific date on which a platform acquires new users from a channel.} 

\textbf{Definition 2 (Retention Day)} \textit{Retention day is used to represent the user retention status of a channel after activation date, typically expressed as day N retention. For example, day 30 retention indicates the user retention status on the 30th day after the activation date.} 

From the channel perspective, we collect the number of activated users for a channel on its activation day and track the retention status of these users across different retention days. We denote the number of activated users for channel $C$ on day $D$ as $\mathbf{A}_{C}^{D}$. The retention ratio is generated daily from the activation date, producing a time series. We refer to this time series as the \textbf{user retention curve}.


Formally, for the time series of activation date $D$, our task is to predict $d$ days starting from day $D+S$, based on $S$ days of observed data. We denote the observed time series as target time series $\mathbf{X}_{tar}=\left\{\mathbf{x}_{1}, \ldots, \mathbf{x}_{S}\right\} \in \mathbb{R}^{S \times 1}$ with $S$ time steps, and predicted time series as $\hat{\mathbf{Y}}=\left\{\mathbf{x}_{S+1}, \ldots, \mathbf{x}_{S+d}\right\} \in \mathbb{R}^{d \times 1}$ with $d$ time steps.

\textbf{Weibull Prior for Retention.} An individual’s retention duration can be regarded as the minimum of multiple latent competing risks. Under this idealized setting, the Fisher–Tippett–Gnedenko theorem~\cite{eaebbc4c-5f6a-3b31-9c38-cfb26bccff39} implies that if a large number of independent and identically distributed risk times possess a finite lower bound and follow a power-law growth near that bound, the limiting distribution of their minimum is exactly the Type III extreme value distribution, i.e., the Weibull distribution~\cite{rinne2008weibull}. Although these strict assumptions may not fully hold in real-world applications, the Weibull form still empirically captures the characteristic decay-then-saturation trend of retention curves.

\textbf{Covariates.} Limited by the length of the look-back window, the model lacks sufficient historical information and struggles to make precise predictions based solely on the input. Therefore, we introduce several covariates to provide additional information to forecast the user retention curve. 

Considering channel heterogeneity, adding identifiers to each channel enables the model to distinguish between different channels and applications. These types of identifiers are static variates that do not change over time, which we denote as $\mathbf{V}_{static}$. 

Meanwhile, on special dates, the user retention curve tends to fluctuate. To enhance the model's ability to learn these distinctive patterns, we introduce date‑related covariates. These covariates are dynamic variates that change over time and are denoted as $\mathbf{V}_{dynamic}$, which mainly fall into two categories: Multi‑calendar Temporal Features and Domain‑specific Events. 

Within Multi‑calendar Temporal Features, we integrate dual‑calendar time characteristics from both the Gregorian and the Chinese lunar calendars. 
For Domain‑specific Events, we explicitly encode key event indicators, including but not limited to statutory holidays, summer vacation. 

\begin{figure*}[t]
\centering
\includegraphics[width=1\textwidth]{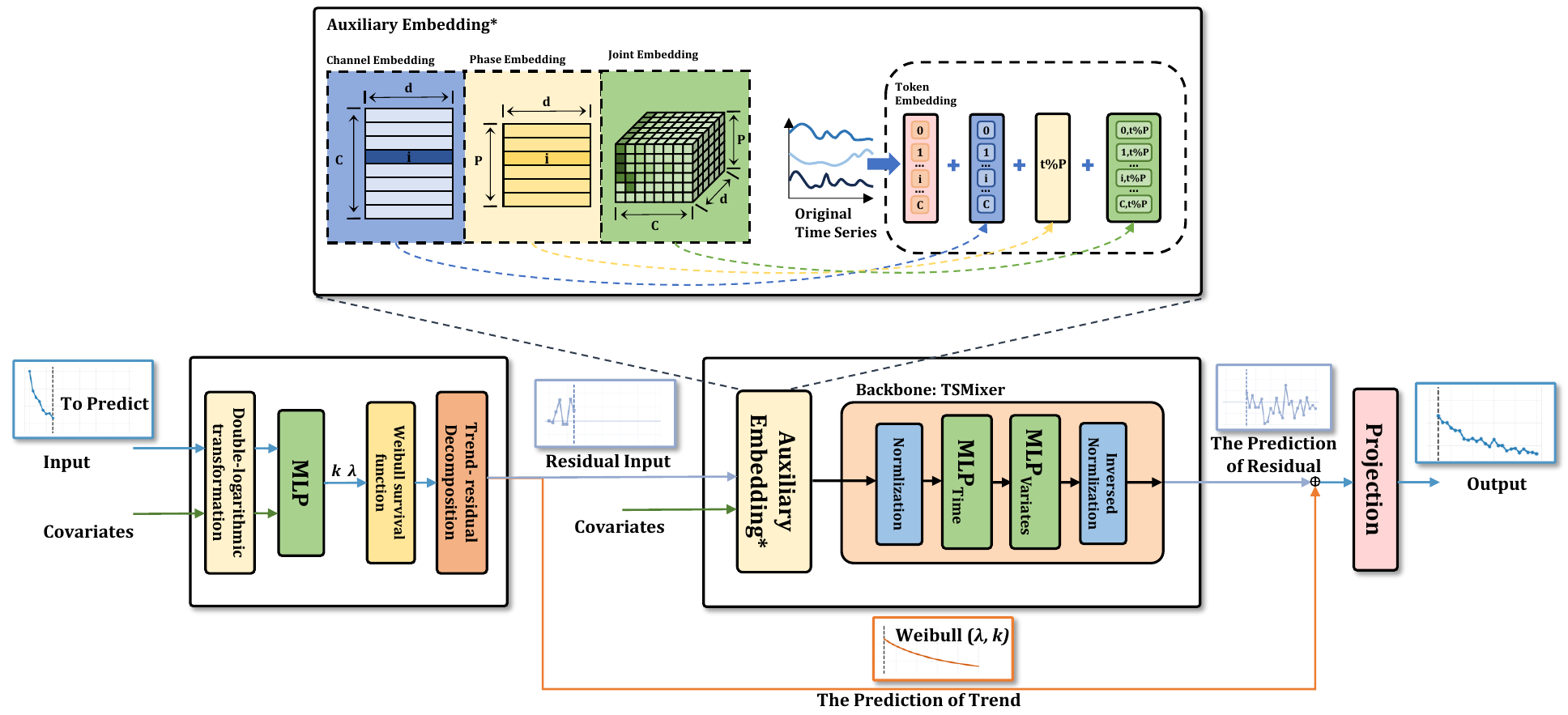}
\setlength{\abovecaptionskip}{0.cm}
\caption{Overall structure of PPDL. Auxiliary Embedding: To enhance the model’s awareness of both different channels and temporal patterns across periodic phases, we design auxiliary embedding, which primarily include channel embedding, phase embedding, and joint embedding. PPDL is a novel framework based on backbone TSMixer, which integrates three key components: Weibull-Prior Trend Extractor, Auxiliary Embedding, and Multiscale Trend-penalized Loss Function.}
\label{Figure 2}
\end{figure*}

\section{Model Design}

In this section, we provide a detailed explanation of the core architecture of PPDL. To address the three major challenges mentioned above, we design three core modules: (1) Weibull-Prior Trend Extractor, (2) Auxiliary Embedding, and (3) Multiscale Trend-penalized Loss Function. In section~\ref {Overall Structure}, we present the overall architecture of PPDL. In section~\ref{Weibull-Prior Trend Extractor}, we apply trend-residual decomposition and introduce Weibull distribution parameterized via MLP to fit trend information based on the observed data. In section~\ref{Auxiliary Embedding}, we design auxiliary embedding to capture inter-channel correlations and periodic temporal patterns in forecasting the residual component. In section~\ref{Multiscale Trend-penalized Loss Function}, to enhance the model's ability to learn trend information, we design a multiscale trend-penalized loss function.

\subsection{Overall Structure} \label{Overall Structure}

We propose PPDL illustrated in Figure~\ref{Figure 2}. Given the short look-back window, we introduce covariates to provide additional information for the model. In terms of the model architecture, we first apply trend-residual decomposition to the input time series. For the trend component, the Weibull distribution is utilized for modeling. For the residual component, we introduce auxiliary embedding, enabling the model to distinguish channel differences while learning time series patterns based on the periodic phase. Finally, we design a multiscale trend-penalized loss function to enhance the model's capability for trend modeling. Since covariates are incorporated, a backbone that supports covariates is required. We adopts the MLP-based model TSMixer~\cite{chen2023tsmixer} as the backbone. The specific details will be discussed in the experiments.

\subsection{Weibull-Prior Trend Extractor} \label{Weibull-Prior Trend Extractor}

 User retention curves exhibit a pronounced global decay-then-saturation trend, where retention drops sharply at the early stage and becomes relatively stable in the long run. This dominant trend often masks fine-grained fluctuations (e.g., weekly periodic patterns), and also introduces strong trend–fluctuation coupling.

To address this issue, the Weibull-Prior trend extractor is designed to explicitly separate the dominant decay trend from the remaining fluctuation signal. 
Concretely, we parameterize the trend as a Weibull survival curve, initialized by fitting a Weibull prior on the look-back observations, and further refined via a gated-MLP conditioned on the look-back observations and covariates.
The network predicts parameter corrections, while a learned confidence gate modulates the magnitude of these updates to balance between the Weibull prior and data-driven flexibility.

Given the input time series $\mathbf{X}_{tar} \in \mathbb{R}^{S}$, we decompose it into a trend component $\mathbf{T} \in \mathbb{R}^{S}$ and a residual component $\mathbf{R} \in \mathbb{R}^{S}$:
\begin{equation}
\mathbf{X}_{tar} = \mathbf{T} + \mathbf{R}.
\end{equation}
Here, $\mathbf{T}$ captures the global decay and long-term saturation, while $\mathbf{R}$ preserves the remaining fluctuations (including potential periodic patterns) for subsequent modeling (see Section~\ref{Auxiliary Embedding}).
We model the trend of each series as a Weibull survival form with a learnable floor
\begin{equation}
\mathbf{T}(t) = f + (1 - f) \cdot \exp \left( -\left( \frac{t}{\lambda} \right)^k \right),
\end{equation}
where $k>0$ and $\lambda>0$ are the shape and scale parameters, and $f\in[0,1]$ denotes the long-run retention floor.

Estimating $(k,\lambda,f)$ from short and noisy histories can be unstable. We therefore adopt a two-stage strategy. 
First, we obtain a robust coarse initialization for $(k,\lambda)$ by applying Theil–Sen regression~\cite{ohlson2015linear} to the log-linearized Weibull form. In this stage, we set $f_{\text{init}}=0$ and fit 
\begin{equation}
y(t)=\ln\big(-\ln(T(t))\big) \approx k\ln t - k\ln\lambda,
\end{equation}
where the Theil–Sen slope and intercept yield $(k_{\text{init}},\lambda_{\text{init}})$.
Second, we refine the parameters using a gated-MLP conditioned on the look-back observations and covariates:
\begin{equation}
(\Delta k, \Delta \lambda, \Delta f, \gamma) = \mathrm{gated\text{-}MLP}(\mathbf{z}, k_{\text{init}}, \lambda_{\text{init}}),
\end{equation}
where $\mathbf{z}=\Phi(\mathbf{X}_{\text{tar}},\mathbf{X}_{\text{cov}})$ summarizes statistics of the look-back observations and covariates.
We interpret $\Delta k$, $\Delta \lambda$, and $\Delta f$ as corrections to the Weibull parameters, and use a confidence gate
$\gamma$ to modulate the magnitude of these corrections.
We then update $k$ with a projected step:
\begin{equation}
k \leftarrow \Pi_{[k_{\min},k_{\max}]}\Big(k_{\text{init}}+\gamma\cdot \Delta k\Big),
\label{eq:update_k}
\end{equation}
which keeps $k$ within $[k_{\min},k_{\max}]$ via projection.
We update $\lambda$ and $f$ in the same form, i.e., $\theta \leftarrow \Pi(\theta_{\text{init}}+\gamma\cdot\Delta\theta)$ for
$\theta\in\\{k,\lambda,f\\}$ (with parameter-specific bounds).

Conditioned on multi-source covariates and look-back observations, the network refines parameters: it keeps them close to the initial Weibull estimates when the pattern is consistent, and increases the correction under irregular dynamics to better match complex multivariate data distributions. 

The calibrated parameters are then used to generate the trend component.

\subsection{Auxiliary Embedding} \label{Auxiliary Embedding}

In time series forecasting, token embedding is primarily achieved through two methods: temporal token embedding~\cite{zhou2021informer,zhou2022fedformer} and variate token embedding~\cite{liu2023itransformer}. Due to the short look-back window, capturing dependencies across channels to supplement predictions with additional information is crucial for improving accuracy. 

We employ variate token embedding with auxiliary embedding to extract effective information from covariates. To enable the model to focus on stably correlated channels and learn similar time series patterns, we introduce channel embeddings, allowing the model to learn the differences and interrelationships among channels. To enhance the model's ability in capturing periodic patterns, we incorporate phase embeddings to attach temporal phase information to the time series, helping the model understand the relative position of data within the phase. Finally, since time series may exhibit different behavioral patterns across phases, we introduce joint embeddings to enable the model to learn spatiotemporal interaction. 

For the given input, the auxiliary embedding $\mathbf{E}_{aux} \in \mathbb{R}^{N \times d}$ is constructed by summing three components:

\begin{equation}
\mathbf{E}_{aux} = \mathbf{W}_{ch} + (\mathbf{1}_N \otimes \mathbf{e}_p) + \text{Reshape}(\mathbf{j}_p).
\end{equation}
\\
where $N$ denotes the number of variates, $d$ denotes the feature dimension. The detailed definitions and functions of each component are as follows:

\textbf{Channel Embedding:} $\mathbf{W}_{ch} \in \mathbb{R}^{N \times d}$ is a learnable static spatial bias. It assigns a unique, time-invariant latent representation vector to each variate token.

\textbf{Phase Embedding}: $\mathbf{e}_p \in \mathbb{R}^{1 \times d}$ is a phase bias obtained via a table lookup operation, where $p$ denotes the phase index to which the current sample belongs. $\mathbf{1}_N \otimes \mathbf{e}_p$ indicates that this global phase information is broadcast to all $N$ channels via the Kronecker product.

\textbf{Joint Embedding}: $\mathbf{j}_p \in \mathbb{R}^{1 \times (N \cdot d)}$ is a spatiotemporal interaction that models phase-specific channel characteristics. It maps the phase index $p$ to a high-dimensional vector, which is then reshaped into a matrix of $\mathbb{R}^{N \times d}$ to interact directly with each channel.

The synergistic effect of three components enables the model to maintain awareness of channel identity while dynamically modulating the influence weights of each covariate based on the current periodic phase. The output of the auxiliary embedding is summed with the output of the token embedding to obtain the result of the data embedding $\mathbf{Z}_{in}$.

\begin{equation}
\mathbf{Z}_{in} = \text{Embedding}(\mathbf{X}) + \mathbf{E}_{aux}.
\end{equation}
\\
where $\mathbf{E}_{aux}$ represents the output of auxiliary embedding. $\text{Embedding}(\mathbf{X})$ represents the token embedding. Specifically, our approach employs variate token embedding.

\subsection{Multiscale Trend-penalized Loss Function} \label{Multiscale Trend-penalized Loss Function}

Considering the pronounced global decay-then-saturation trend of user retention curves, we introduce the multiscale trend-penalized loss function to augment the Weibull distribution in modeling trend information. Building upon the Mean Squared Error (MSE), this function introduces a multi-scale difference term aimed at simultaneously constraining both the accuracy of predicted values and the consistency of their changing trends. The loss function $\mathcal{L}_{total}$ is defined as a weighted combination of the base MSE loss and the trend-penalized term:
\begin{equation}
\mathcal{L}_{total} = (1 - \alpha) \mathcal{L}_{mse} + \alpha \mathcal{L}_{trend}.
\end{equation}
\\
where $\alpha$ is a hyperparameter that controls the strength of the trend constraint. In order to capture trend characteristics across different time spans, we introduce a set $\mathcal{T} = \{k_1, k_2, \dots\}$ to represent different scale factors (e.g., week, month). For each scale $k \in \mathcal{T}$, we compute the scale-wise difference loss. The formal expression is as follows:

\begin{equation}
\mathcal{L}_{trend} = \frac{1}{|\mathcal{T}|} \sum_{k \in \mathcal{T}} \text{MSE}\left( \Delta^k \hat{\mathbf{Y}}, \Delta^k \mathbf{Y} \right)
\end{equation}

where $\Delta^k$ denotes the difference operator with scale $k$. $\hat{\mathbf{Y}}$ represents the forecasting result, and $\mathbf{Y}$ represents the ground truth.

\section{Experiments}

\label{sec:experiments}
We evaluate our framework on datasets from three applications in a short-video platform, each with user bases in the hundreds of millions. 
Our experiments demonstrate that PPDL can effectively address the challenges of long-horizon user retention forecasting under limited observation windows.
\subsection{Datasets}
\begin{figure}[t]
    \centering
    \includegraphics[width=1\columnwidth]{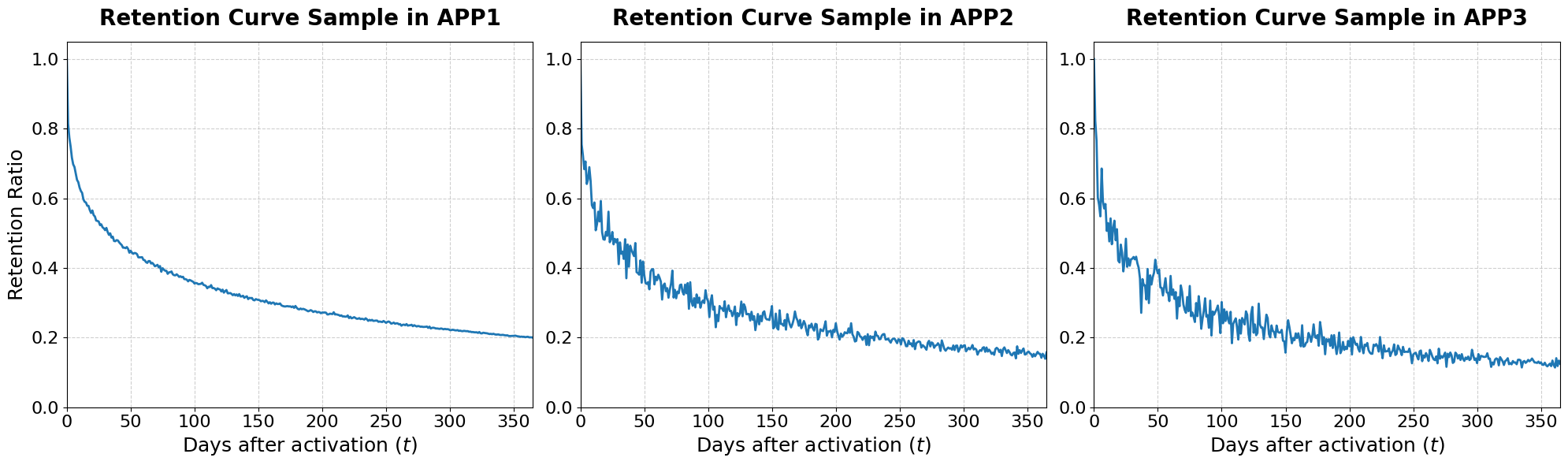}
    \setlength{\abovecaptionskip}{0.cm}
    \caption{ App1 (Large-scale): Features a massive user base with relatively uniform demographic distributions. App2 (Medium-scale): Represents a moderate-sized user base with slight distribution skews. App3 (Small-scale): Contains a smaller user bases with notable distribution shifts.}
    \label{Figure 3}
\end{figure}

As show in Figure~\ref{Figure 3}, there are three different applications. Representing diverse business scales and user distribution characteristics:
\begin{itemize}
\item App1 (Large-scale): Features a massive user base with relatively uniform demographic distributions. The large sample size and balanced user composition provide high-quality data with strong signals, making retention trend estimation more stable and reliable.
\item App2 (Medium-scale): Represents a moderate-sized user base with slight distribution skews. This dataset reflects typical conditions in production environments, where user demographics exhibit mild imbalances while maintaining reasonable data quality.
\item App3 (Small-scale): Contains a smaller user sample with notable distribution shifts. The limited scale and skewed user demographics introduce higher volatility and noise, presenting a more challenging scenario for robust trend estimation.
\end{itemize}

We evaluate user retention ratio forecasting across three applications with 30+ channels per app on average, capturing retention patterns from user bases averaging hundreds of millions per application. This diverse evaluation landscape allows us to assess our model's performance under varying conditions of user scale, demographic distribution, and data quality, providing comprehensive insights into its robustness and generalization capabilities.

\subsection{Evaluation Metrics}
To evaluate the accuracy of predicting long-term user retention, we employ the {Weighted MAPE of Cumulative Retention ($\mathrm{MAPE}_a$)} as the primary metric. This metric measures the forecasting error for the {cumulative number of retained users} from day 30 to 365. It directly aligns with the accuracy of long-term user growth and budget allocation strategies, providing more actionable business insight than daily pointwise error metrics.

It is formulated as follows:
\begin{equation}
\text{MAPE}_a = \frac{\sum_{C, D} \mathbf{A}_{C}^{D} \cdot \mathrm{MAPE}(\widehat{CR}_{C}^{D}, CR_{C}^{D})}{\sum_{C, D} \mathbf{A}_{C}^{D}},
\label{eq:metric_w_mape_c}
\end{equation}
where \( \widehat{CR}_{C}^{D} = \sum_{i=30}^{365} \hat{\mathbf{R}}_{C}^{D}[i] \) and \( CR_{C}^{D} = \sum_{i=30}^{365} \mathbf{R}_{C}^{D}[i] \) denote the predicted and ground-truth cumulative retention ratio for channel \( C \) on activation date \( D \), respectively. \( \mathbf{A}_{C}^{D} \) is the number of activated users, which serves as the weight to ensure that channels with larger user volumes contribute more to the overall metric, thereby reflecting the global business impact more accurately.


\subsection{Main Results}
\label{subsec:main_results}

We conduct a comparative evaluation of our proposed PPDL framework against the primary TSMixer across three production applications, and also compare it with the currently online-deployed statistical model. Table~\ref{tab:main_results} reports the performance using the weighted $\mathrm{MAPE}_{a}$ metric.

\begin{table}[htbp]
\caption{Performance comparison on three production datasets. Error is measured by the weighted $\mathrm{MAPE}{a}$ (\%). The $\Delta$ $\mathrm{MAPE}{a}$ (Abs.) and $\Delta$ $\mathrm{MAPE}_{a}$ (Rel. \%) metrics represent improvements relative to TSMixer. Bold denotes the best performance.}
\centering
\resizebox{\columnwidth}{!}{
\begin{tabular}{cccccc}
\toprule
Application & \textbf{PPDL (Ours)} & TSMixer & Statistical model & $\Delta$ $\mathrm{MAPE}_{a}$(Abs.)  & $\Delta$ $\mathrm{MAPE}_{a}$(Rel. \%)   \\
\midrule
App1 & \textbf{3.6\%} & 5.7\%  & 4.9\% &  -2.1\% & -36.8\% \\
App2 & \textbf{5.0\%} &  5.9\% & 6.4\% & -0.9\% & -15.3\% \\
App3 & \textbf{6.6\%} &  8.4\%& 7.7\% & -1.8\% & -21.4\% \\
\midrule
\textbf{Avg.} & \textbf{5.07\%} & 6.67\% & 6.33\% & -1.60\% & -24.50\% \\
\bottomrule
\end{tabular}
}

\label{tab:main_results}
\end{table}

As shown in Table~\ref{tab:main_results}, the PPDL framework consistently outperforms the primary TSMixer in all scenarios. On average, PPDL achieves a $\mathrm{MAPE}_{a}$ of 5.07\%, representing a {24.5\% relative reduction} (1.60\% in absolute $\Delta$) compared to the primary TSMixer (6.67\%). The most significant improvement is achieved in App1, with a relative error reduction of 36.8\%. This demonstrates that the inductive bias provided by the Weibull-prior trend extractor effectively addresses information scarcity in the observation window.

The performance gains are also substantial in App3, which faces more challenging data quality and smaller user scales. PPDL achieves a 21.4\% relative reduction in App3, compared to 15.3\% in App2. This highlights that PPDL is particularly effective when the signal-to-noise ratio is low, as the explicitly modeled Weibull trend provides a robust backbone that prevents the neural network from overfitting to short-term fluctuations.

Compared to the currently online-deployed statistical model, PPDL also achieves consistent and substantial improvements. On average, PPDL reduces the $\mathrm{MAPE}_{a}$ from 6.33\% to 5.07\%, corresponding to a relative reduction of 19.9\% (1.26\% in absolute $\Delta$). The improvements are most pronounced in App1 (26.5\%) and App2 (21.9\%), while App3 still benefits from a 14.3\% relative reduction. These results highlight that the deep learning-based PPDL framework, augmented with physical priors, not only surpasses generic neural baselines but also significantly outperforms domain-specific statistical models that rely on static parametric forms and expert heuristics.

\subsection{Ablation Studies}
\label{subsec:ablation}
To systematically assess the individual contribution of each architectural component, we conduct a series of ablation studies. Each experiment isolates specific modules while maintaining the backbone TSMixer architecture for all other components.

\subsubsection{Impact of Trend-residual Decomposition}
\leavevmode\newline
This section evaluates the individual contribution of the trend-residual decomposition module. Table~\ref{tab:abl_trend} reveals that this component delivers more substantial improvements than the auxiliary embedding module alone, achieving an average relative error reduction of 19.23\%. The effect is particularly pronounced in App1, which exhibits a 29.8\% relative improvement. These findings underscore the critical importance of explicitly disentangling trend components from seasonal and residual variations. By separately modeling these distinct temporal patterns, the architecture can more effectively capture the underlying structural dynamics of the time series.

\begin{table}[ht]
\caption{Ablation study on the trend-residual decomposition module. The results indicate an average relative reduction of 19.23\% in MAPE, confirming that the decomposition module effectively enhances the accuracy of forecasting for complex decay trends.}
\centering
\resizebox{\columnwidth}{!}{
\begin{tabular}{ccccc}
\toprule
Application & \begin{tabular}[c]{@{}c@{}} \textbf{PPDL -Aux. Emb.}\\ \textbf{-Mult. Trend. Loss} \end{tabular} & TSMixer & $\Delta$ $\mathrm{MAPE}_{a}$(Abs.) & $\Delta$ $\mathrm{MAPE}_{a}$(Rel. \%) \\
\midrule
App1 & \textbf{4.0\%} & 5.7\% & -1.7\% & -29.8\% \\
App2 &  \textbf{5.1\%}& 5.9\% & -0.8\% & -13.6\% \\
App3 &  \textbf{7.2\%} &8.4\% & -1.2\% & -14.3\% \\
\midrule
\textbf{Avg.} & \textbf{5.43\%}& 6.67\%  & -1.23\% & -19.23\% \\
\bottomrule
\end{tabular}
}

\label{tab:abl_trend}
\end{table}

\subsubsection{Impact of the Auxiliary Embedding Module}
\leavevmode\newline
This section evaluates the individual contribution of incorporating the auxiliary embedding module into the TSMixer architecture. This module encodes channel identity, temporal phase, and their joint interactions to capture inter-channel correlations and periodic patterns. As shown in Table~\ref{tab:abl_aux}, the auxiliary embedding module yields consistent improvements across all three applications, reducing the average $\mathrm{MAPE}_{a}$ from 6.67\% to 6.00\%. This corresponds to an absolute reduction of 0.67\% and a relative improvement of 9.97\%. Notably, the gain is most significant in App1, which exhibits a 15.8\% relative reduction. These results indicate that the enriched feature representation provided by the auxiliary embedding effectively stabilizes forecasting results by leveraging complementary contextual information across channels and time.

\begin{table}[ht]
\caption{Ablation study on the auxiliary embedding module. The inclusion of auxiliary embeddings leads to an average relative reduction of 9.97\% in MAPE, demonstrating the effectiveness of incorporating shared information across different applications.}
\centering
\resizebox{\columnwidth}{!}{
\begin{tabular}{ccccc}
\toprule
{Application} & \begin{tabular}[c]{@{}c@{}} \textbf{PPDL -Trend}\\ \textbf{-Mult. Trend. Loss} \end{tabular}  & TSMixer& {$\Delta$ $\mathrm{MAPE}_{a}$(Abs.)} & {$\Delta$ $\mathrm{MAPE}_{a}$(Rel. \%)} \\
\midrule
App1 & \textbf{4.8\%} & 5.7\% & -0.9\% & -15.8\% \\
App2 & \textbf{5.7\%} & 5.9\% & -0.2\% & -3.4\% \\
App3 & \textbf{7.5\%}  & 8.4\%& -0.9\% & -10.7\% \\
\midrule
\textbf{Avg.} &  \textbf{6.00\%}& {6.67\%} & {-0.67\%} & {-9.97\%} \\
\bottomrule
\end{tabular}
}

\label{tab:abl_aux}
\end{table}

\subsubsection{Combined Effect of Trend-residual Decomposition and Auxiliary Embedding}
\leavevmode\newline
\begin{table}[ht]
\caption{Ablation study on the combined effect of trend-residual decomposition and auxiliary Embedding module. The integrated configuration achieves an average relative reduction of 22.53\% in MAPE, highlighting the importance of these components for robust trend modeling.}
\centering
\resizebox{\columnwidth}{!}{
\begin{tabular}{ccccc}
\toprule
Application& \textbf{PPDL -Mult. Trend. Loss} &  TSMixer & $\Delta$ $\mathrm{MAPE}_{a}$(Abs.) & $\Delta$ $\mathrm{MAPE}_{a}$(Rel. \%) \\
\midrule
App1 &  \textbf{3.8\%} & 5.7\%& -1.9\% & -33.3\% \\
App2 & \textbf{5.0\%}&   5.9\%& -0.9\% & -15.3\% \\
App3 & \textbf{6.8\%} & 8.4\% & -1.6\% & -19.0\% \\
\midrule
\textbf{Avg.} &\textbf{5.20}\%  & {6.67}\% & -1.47\% & -22.53\% \\
\bottomrule
\end{tabular}
}

\label{tab:abl_trend_ema}
\end{table}

This section evaluates the combined effect of the trend-residual decomposition and auxiliary Embedding module. As shown in Table~\ref{tab:abl_trend_ema}, integrating both modules reduces the average $\mathrm{MAPE}_{a}$ by 1.47\%, corresponding to a relative reduction of 22.53\%. The improvement is most significant for App1, which exhibits a 33.3\% relative
reduction. We further compare the combined configuration (trend-residual decomposition + auxiliary embedding) against each individual module. As shown in Table~\ref{tab:abl_trend_ema_2}, the combined modules achieve the lowest average $\mathrm{MAPE}_{a}$ (5.20\%), outperforming the trend-only (5.43\%) and auxiliary-only (6.00\%) variants. The relative improvements of the combined modules over the trend-only and auxiliary-only variants are 4.20\% and 14.13\% on average, respectively. Notably, the improvement over the auxiliary-only variant is substantially larger than over the trend-only variant across all three applications, indicating that the trend-residual decomposition provides a fundamental contribution to long-term forecasting accuracy. These results indicate that the combination of two modules addresses complementary aspects of the forecasting task: the trend-residual decomposition provides a physically grounded long-term prior, while the auxiliary embedding refines short-term and medium-term fluctuations by leveraging cross-channel and temporal context. Their synergy leads to robust and accurate forecasting results, especially when the signal-to-noise ratio is low (e.g., App3).

\begin{table}[ht]
\caption{Comparison of the combined modules against each single module. +Trend+Aux. Emb. denotes the result of the combination of the trend-residual decomposition and auxiliary Embedding module, +Trend and +Aux. Emb. denote the results of the trend-only and auxiliary-only variants, respectively. Rel. +Trend and Rel. +Aux. Emb. denote the relative improvement of the combined modules over the respective single module.}
\centering
\resizebox{\columnwidth}{!}{
\begin{tabular}{cccccc}
\toprule
Application& \textbf{+Trend+Aux. Emb.} &  +Trend & +Aux. Emb. &Rel. +Trend   &Rel. +Aux. Emb.\\
\midrule
App1 &  \textbf{3.8}\%& 4.0\% & 4.8\% & -5.0\% & -20.8\% \\
App2 &  \textbf{5.0}\%& 5.1\% & 5.7\% & -2.0\% & -12.3\% \\
App3 & \textbf{6.8}\% & 7.2\% & 7.5\% & -5.6\% & -9.3\% \\
\midrule
\textbf{Avg.} & {\textbf{5.20}}\% & 5.43\% & 6.00\% & -4.20\% & -14.13\%\\
\bottomrule
\end{tabular}
}

\label{tab:abl_trend_ema_2}
\end{table}

\subsubsection{Impact of Multiscale Trend-penalized Loss}
\leavevmode\newline
This section evaluates the individual contribution of the Multiscale Trend-penalized Loss. Table~\ref{tab:abl_loss} reveals that this component achieves an average relative error reduction of 2.23\%. The effect is particularly pronounced in App1, which exhibits a 3.4\% relative improvement. These findings demonstrate that the proposed loss function is more robust to data irregularities, effectively reducing the interference of noise. By mitigating the impact of random fluctuations, the architecture can more accurately capture the true underlying structural dynamics of the time series.

\begin{table}[htbp]
\caption{Ablation study on Multiscale Trend-penalized Loss. The results show an average relative reduction of 2.23\% in MAPE, confirming the effectiveness of the multiscale trend-penalized mechanism.}
\centering
\resizebox{\columnwidth}{!}{
\begin{tabular}{ccccc}
\toprule
Application & \textbf{PPDL} & PPDL -Mult. Trend. Loss & $\Delta$ $\mathrm{MAPE}_{a}$(Abs.) & $\Delta$ $\mathrm{MAPE}_{a}$(Rel. \%) \\
\midrule
App1 & \textbf{3.6\%} & 3.8\% & -0.2\% & -3.4\% \\
App2 & \textbf{5.0\%} & 5.0\% & -0.0\% & -0.0\% \\
App3 & \textbf{6.6\%} & 6.8\% & -0.2\% & -3.3\% \\
\midrule
\textbf{Avg.} & \textbf{5.07\%} & 5.20\% & -0.13\% & -2.23\% \\
\bottomrule
\end{tabular}
}

\label{tab:abl_loss}
\end{table}

\subsubsection{Comparison with Online Statistical Model}
\label{subsec:online_comparison}
\leavevmode\newline
We compare our approach with the previously deployed online model, which is a statistical framework with expert knowledge. It can achieve reliable forecasting performance and has been in different productions for many years. Table~\ref{tab:abl_with_static_model} reveals that our approach delivers substantial improvements, achieving an average relative error reduction of 20.90\%. The effect is particularly pronounced in App1, which exhibits a 26.5\% relative improvement. These findings underscore the advantage of our model's ability to perform personalized modeling for distinct applications while maintaining robustness. In contrast, the statistical baseline is limited by its parameter capacity, which restricts its ability to adapt to individual app characteristics and makes it more susceptible to bias.


\begin{table}[ht]
\caption{Comparison with Online Statistical Model. The proposed approach achieves an average reduction of 20.90\% in relative error. These results highlight the advantage of personalized modeling over the statistical baseline.}
\centering
\resizebox{\columnwidth}{!}{
\begin{tabular}{ccccc}
\toprule
Application & \textbf{PPDL} & Statistical model & $\Delta$ $\mathrm{MAPE}_{a}$(Abs.) & $\Delta$ $\mathrm{MAPE}_{a}$(Rel. \%) \\
\midrule
App1 & \textbf{3.6\%} & 4.9\% & -1.3\% & -26.5\% \\
App2 & \textbf{5.0\%} & 6.4\% & -1.4\% & -21.9\% \\
App3 & \textbf{6.6\%} & 7.7\% & -1.1\% & -14.3\% \\
\midrule
\textbf{Avg.} & \textbf{5.07\%} & 6.33\% & -1.27\% & -20.90\% \\
\bottomrule
\end{tabular}
}

\label{tab:abl_with_static_model}
\end{table}

\begin{figure*}[ht]
\centering
\begin{subfigure}[b]{0.24\textwidth}
\centering
\includegraphics[width=\textwidth, trim=0 0 43cm 0, clip]{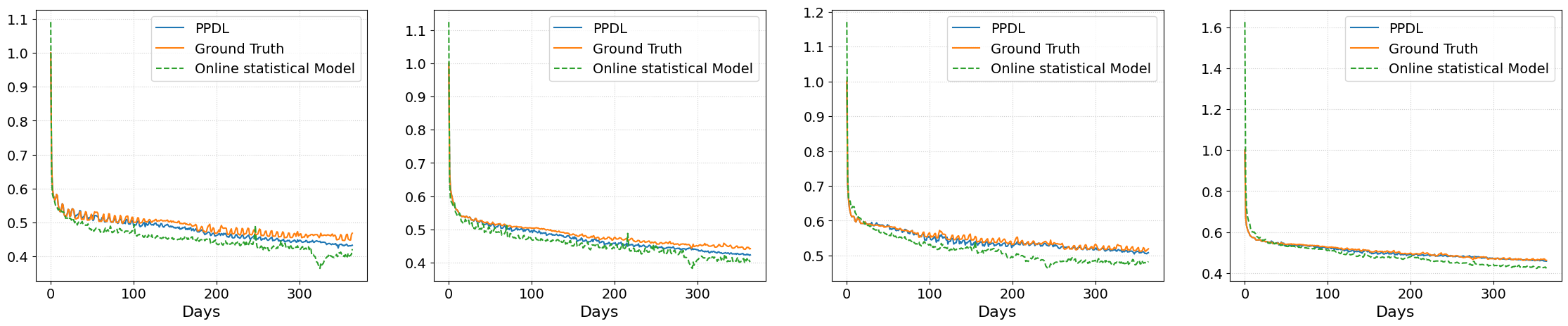}
\caption{Date\_1}
\end{subfigure}
\hfill
\begin{subfigure}[b]{0.24\textwidth}
\centering
\includegraphics[width=\textwidth, trim=14.3cm 0 28.6cm 0, clip]{fig_exp/fig_1.png}
\caption{Date\_2}
\end{subfigure}
\hfill
\begin{subfigure}[b]{0.24\textwidth}
\centering
\includegraphics[width=\textwidth, trim=28.6cm 0 14.3cm 0, clip]{fig_exp/fig_1.png}
\caption{Date\_3}
\end{subfigure}
\hfill
\begin{subfigure}[b]{0.24\textwidth}
\centering
\includegraphics[width=\textwidth, trim=43cm 0 0 0, clip]{fig_exp/fig_1.png}
\caption{Date\_4}
\end{subfigure}

\caption{Comparison of PPDL and the online model on four APP1 cases: (a)–(c) one channel with three activation dates; (d) another channel on a different activation date. The blue, green, and orange curves represent PPDL, online, and the ground truth. X-axis: retention day; Y-axis: retention rate. PPDL follows the trend more robustly, with less bias and fewer fluctuations.}
\label{fig:mff_prediction_comparison}
\end{figure*}

\begin{figure*}[htbp]
  \centering
\begin{subfigure}[b]{0.24\textwidth}
\centering
\includegraphics[width=\textwidth, trim=0 0 43cm 0, clip]{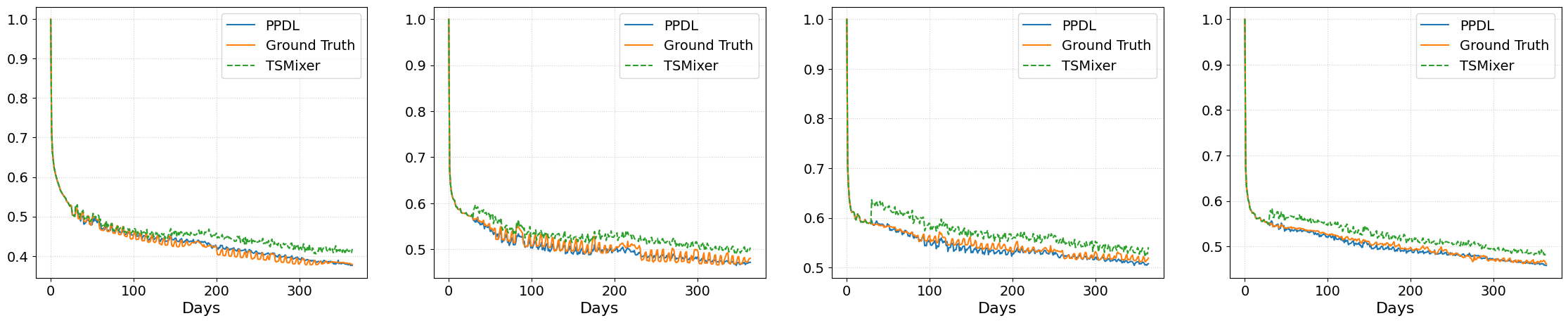}
\caption{Date\_1}
\end{subfigure}
\hfill
\begin{subfigure}[b]{0.24\textwidth}
\centering
\includegraphics[width=\textwidth, trim=14.3cm 0 28.6cm 0, clip]{fig_exp/fig_tsmixer.png}
\caption{Date\_2}
\end{subfigure}
\hfill
\begin{subfigure}[b]{0.24\textwidth}
\centering
\includegraphics[width=\textwidth, trim=28.6cm 0 14.3cm 0, clip]{fig_exp/fig_tsmixer.png}
\caption{Date\_3}
\end{subfigure}
\hfill
\begin{subfigure}[b]{0.24\textwidth}
\centering
\includegraphics[width=\textwidth, trim=43cm 0 0 0, clip]{fig_exp/fig_tsmixer.png}
\caption{Date\_4}
\end{subfigure}

  \caption{Comparison of forecasting results between PPDL and the primary TSMixer. The figure illustrates four representative cases, each corresponding to the retention rate forecast for a combination of application, primary channel, and secondary channel under specific activation dates. The blue, green, and orange curves represent PPDL, primary TSMixer, and the ground truth. X-axis: retention day; Y-axis: retention rate.}
  
  \label{fig:mmf_vs_tesmixer}
\end{figure*}

\subsection{Generalization Analysis}
To comprehensively evaluate the generality of the PPDL framework, we select three representative backbones that cover a wide range of design philosophies in time series forecasting. TSMixer~\cite{chen2023tsmixer} is a lightweight, MLP-based model that achieves competitive performance with low computational cost, making it an ideal baseline for efficiency-oriented scenarios.  iTransformer~\cite{liu2023itransformer}, by inverting the conventional Transformer and treating each variate as a token, leverages self-attention mechanisms to capture cross-channel dependencies, demonstrating strong multivariate forecasting capabilities. TFT~\cite{LIM20211748} is specifically designed for interpretable multi-step forecasting, integrating LSTMs, self-attention, and variate selection networks to handle heterogeneous inputs. By covering MLP-based, Transformer-based, and hybrid architectures, this selection enables us to rigorously test whether PPDL can consistently improve forecasting accuracy across different backbone architectures.

\begin{table}[htbp]
\caption{Performance comparison of different backbone architectures under the PPDL framework. The iTr. represent primary iTransformer. PPDL(T), PPDL(I), and PPDL(F) denote the PPDL framework with TSMixer, iTransformer, and TFT backbones, respectively. Numbers in parentheses indicate the relative change in $\mathrm{MAPE}_{a}$ compared to the corresponding primary backbone. Bold denotes the best performance.}
\centering
\resizebox{\columnwidth}{!}{
\begin{tabular}{ccccccc}
\toprule
Application & TSMixer & iTr. & TFT & PPDL(T) & PPDL(I) &PPDL(F)\\
\midrule
App1 & 5.7\% &  4.7\% & 5.0\% &\textbf{3.6\%(-36.8\%)}  & 4.5\%(-4.2\%)& 4.5\%(-10.0\%)\\
App2 & 5.9\% &  7.9\% & 5.0\% & 5.0\%(-15.2\%) & 5.4\%(-31.6\%)& \textbf{4.4\%(-12.0\%)}\\
App3 & 8.4\% &  19.9\% & 15.7\% & \textbf{6.6\%(-21.4\%)}& 13.2\%(-33.7\%)& 10.4\%(-33.8\%)\\ 
\midrule
\textbf{Avg.} & 6.67\% & 10.83\% & 8.57\% &\textbf{5.07\%(-23.99\%)}& 7.70\%(-28.90\%)& 6.43\%(-24.97\%)\\
\bottomrule
\end{tabular}
}

\label{tab:robustness}
\end{table}

As shown in Table~\ref{tab:robustness}, we evaluate the generality of the PPDL framework under different backbones. For comparison, we also report the forecasting performance of the primary backbones, i.e., TSMixer, iTransformer, and TFT. Experiments show that the original PPDL (with TSMixer as the backbone) achieves the best average $\mathrm{MAPE}_{a}$ (5.07\%) across all three applications, significantly outperforming the primary TSMixer, with a relative error reduction of 23.99\%, which again validates the effectiveness of the PPDL modules. When the backbone is replaced with iTransformer and TFT, the PPDL framework still consistently improves forecasting accuracy: PPDL(I) achieves an average $\mathrm{MAPE}_{a}$ of 7.7\%, which is 28.90\% lower than that of primary iTransformer; PPDL(F) achieves an average $\mathrm{MAPE}_{a}$ of 6.43\%, 24.97\% lower than that of primary TFT. It should be noted that the primary iTransformer and the primary TFT perform significantly worse in App3 ($\mathrm{MAPE}_{a}$ as high as 19.9\% and 15.7\%, respectively), while PPDL(I) and PPDL(F) reduce them to 13.2\% and 10.4\%, respectively. This indicates that the trend residual decomposition and physical priors module in the PPDL framework effectively mitigates overfitting and noise amplification in complex data, thus improving adaptability to different backbones. In summary, the PPDL framework not only achieves the best performance with its native backbone (TSMixer), but also significantly improves the forecasting robustness of other backbone architectures, demonstrating great generality.

\subsection{Visualization Analysis}

We conduct a visualization analysis of PPDL's predictive behavior along two dimensions. On one hand, we compare PPDL with the online statistical model, focusing on its capability for long-term trend correction and noise suppression. On the other hand, we compare PPDL with primary TSMixer to isolate the independent contributions of each module, with emphasis on evaluating the model's ability to capture turning-point timing and plateau shape.

\subsubsection{Long-Term Trend Correction and Noise Suppression vs. the Online Model}
\leavevmode\newline
We show several production channels and compare PPDL against the online statistical model(Figure~\ref{fig:mff_prediction_comparison}). 
PPDL performs notably better in the tail: the online model drifts below the ground truth, producing overly pessimistic long-horizon estimates, whereas PPDL stays closer and mitigates bias accumulation. 
PPDL also reduces spurious high-frequency wiggles, yielding smoother, decision-stable trajectories for downstream budget allocation when the ground truth evolves smoothly.

\subsubsection{Turning-Point and Plateau-Phase Modeling vs. the Backbone (TSMixer)}
\leavevmode\newline
To isolate the contribution of our modules beyond the TSMixer backbone, we compare PPDL with TSMixer on four channel–date pairs (Figure~\ref{fig:mmf_vs_tesmixer}).
TSMixer tends to mis-handle the regime change from early steep decay to the mid/late plateau: its predictions often lag around the turning point and may overshoot the ground truth in the later stage, indicating an inaccurate saturation level. 
In contrast, PPDL better captures the turning-point timing and the subsequent plateau shape, producing curves that track the transition more faithfully. 

\section{Conclusion}

In this paper, we study the challenge of early channel-level user retention ratio forecasting in the scenario of real-world short-video platform. To address the challenges of channel heterogeneity, pronounced global decay-then-saturation trend, and short look-back windows, we design PPDL, a novel framework which integrates three key components:  Weibull-Prior trend extractor, auxiliary embedding, and multiscale trend-penalized loss function. Weibull-Prior trend extractor adapts a gated MLP model for channel-specific parameter estimating. Auxiliary embedding is applied in the TSMixer backbone. Multiscale trend-penalized loss function contributes to handling the trends better. To the best of our knowledge, this is the first work to use deep learning-based time series forecasting for channel-level user retention ratio forecasting. Extensive experiments on real-world industrial datasets demonstrate that PPDL substantially outperforms the online solutions, validating the efficacy of channel-level user retention ratio forecasting.

\section{Related Work}

\subsection{User-level Retention Ratio Forecasting}
In user-level retention ratio forecasting, early approaches are dominated by statistical and probabilistic models. Among the most representative is the Pareto/NBD model~\cite{7d5d69bc-539a-3e6b-bdf0-d509ffc75c8b}, which lays the foundation for subsequent research by modeling user transaction rates with a Gamma distribution and user dropout behavior with a Pareto distribution within a continuous-time framework. 
The BG/BB~\cite{e705ee59-5241-376a-b62c-84fbc1b2e32b} and BdW~\cite{FADER20181} models further extend the framework to contractual settings and introduce duration-dependent structures, allowing retention probabilities to vary over time. However, these statistical approaches are constrained by strict parametric assumptions and the limited capacity to incorporate exogenous features.

With the development of deep learning and the accumulation of fine-grained user behavioral data, graph neural networks and transformer architectures have brought new perspectives to this problem. The hierarchical user churn prediction model~\cite{10606414} based on graph attention convolutional neural networks improves churn prediction accuracy for telecom operators by modeling the graph structure of user interaction behaviors. The approach~\cite{10600684} that combines graph convolutional networks with label propagation techniques has demonstrated the contribution of social activity information to prediction accuracy in player churn prediction for massively multiplayer online role-playing games. Meanwhile, Transformer architectures have also been introduced into churn prediction tasks. Hoang et al.~\cite{10730924} apply the FT-Transformer architecture to early churn prediction in freemium mobile games, enabling early warning when player behavior data is sparse.

As businesses increasingly demand interpretability and personalized intervention capabilities, researchers have begun to explore hybrid approaches that go beyond the pure prediction paradigm.  The SEC framework~\cite{10.1145/3746252.3761505} introduces the Stratified Expert Cloning imitation learning framework, which learns robust policies from the interaction data of highly retained users. This approach achieved a cumulative increase in active days of approximately 0.1\% on two short-video platforms with hundreds of millions of daily active users, translating to an increment of over 200,000 daily active users. 

Despite the significant advances in user-level retention ratio forecasting approaches in recent years, a fundamental limitation they commonly face is that, within a short observation window, the behavioral history of new users is extremely sparse. This makes user-level feature vectors insufficient to support reliable single-user predictions. Moreover, most deep learning models treat churn as a classification problem at a fixed time boundary, rather than modeling churn propensity as a time-continuous function of survival probability, thereby severing the dynamic evolution process of churn tendency.

\subsection{Channel-level User Retention Ratio Forecasting}

Unlike user-level retention ratio forecasting, which has been widely studied and deployed across various industries, the problem of forecasting future retention curves for the same cohort of users at the channel level has received far less attention. In channel-level user retention ratio forecasting, existing approaches~\cite{cherkashin2024practical} are relatively scarce and predominantly rely on statistical methods. Statistical methods can leverage physical priors to fit curves with corresponding characteristics. However, when confronted with large-scale, heterogeneous data in real-world industrial scenarios, their forecasting accuracy is limited, and they struggle to capture complex temporal fluctuation patterns. 


However, deep learning methods, particularly Transformer-based time series models~\cite{vaswani2017attention,chen2024pathformer,nie2022time}, have demonstrated remarkable capabilities in learning long-range dependencies and temporal fluctuation patterns across a variety of forecasting tasks. Models such as Autoformer~\cite{wu2021autoformer}, Crossformer~\cite{zhang2023crossformer}, and iTransformer~\cite{liu2023itransformer} have introduced innovative architectural designs to address the limitations of the original Transformer in time series modeling. Despite their success on time series benchmarks, applying them to channel-level user retention ratio forecasting still poses considerable challenges. First, the quadratic complexity of the self-attention mechanism~\cite{zeng2023transformers} with respect to sequence length incurs high computational cost when forecasting hundreds of time steps. Second, these models are highly sensitive to noise and irregular fluctuations, which are prevalent in short-window, sparse-channel data. With only 30 days of observed data, deep learning models can easily overfit the high-frequency variations within the early retention window, leading to severe extrapolation errors for the remaining 335 days. Therefore, integrating physical priors with deep learning—by modeling the trend and residual components separately based on trend-residual decomposition—can combine the strengths of both methods to improve forecasting performance.

\clearpage
\bibliographystyle{IEEEtran}
\bibliography{IEEEexample}

\end{document}